\documentclass[11pt]{article}
\usepackage[margin=1in]{geometry}
\usepackage[authoryear,round]{natbib}

\usepackage{url,hyperref,lineno,microtype, subcaption}
\usepackage[onehalfspacing]{setspace}

\usepackage{graphicx}
\usepackage{float}

\usepackage{todonotes}
\usepackage{algorithm}
\usepackage{algpseudocode}
\usepackage{algorithmicx}
\usepackage{algorithm} 
\usepackage{listings}
\usepackage{booktabs}   
\usepackage{multirow}   
\usepackage{makecell}   
\usepackage[table]{xcolor} 

\usepackage{amsmath}
\usepackage{amssymb}
\usepackage{url,hyperref,cleveref}
\usepackage{bm}

\algdef{SE}[SUBALG]{Indent}{EndIndent}{}{\algorithmicend\ }%
\algtext*{Indent}
\algtext*{EndIndent}

\begin{document}
\onecolumn





\title{Federated Deep Learning for Privacy-Preserving Cardiovascular Disease Risk Prediction}

\author{Hyunho Mo$^{1,*}$, Djura Smits$^{2}$, Mahlet A. Birhanu$^{1}$, Maarten J.G. Leening$^{1,3,4}$,\\
Daniel Bos$^{1,3}$, Pim van der Harst$^{5}$, Esther E. Bron$^{1}$}

\date{}

\maketitle


\begin{center}
\begin{minipage}{0.9\textwidth}
\footnotesize\centering
$^{1}$Department of Radiology \& Nuclear Medicine, Erasmus MC University Medical Center Rotterdam, Dr. Molewaterplein 40, Rotterdam, 3015 GD, The Netherlands\\
$^{2}$Netherlands eScience Center, Matrix THREE, Science Park 402, Amsterdam, 1098 XH, The Netherlands\\
$^{3}$Department of Epidemiology, Erasmus MC University Medical Center Rotterdam, Dr. Molewaterplein 40, Rotterdam, 3015 GD, The Netherlands\\
$^{4}$Department of Cardiology, Erasmus MC University Medical Center Rotterdam, Dr. Molewaterplein 40, Rotterdam, 3015 GD, The Netherlands\\
$^{5}$Department of Cardiology, University Medical Center Utrecht, Heidelberglaan 100, Utrecht, 3584 CX, The Netherlands\\[3pt]
$^{*}$Correspondence: Hyunho Mo, \texttt{h.mo@erasmusmc.nl}
\end{minipage}
\end{center}

\vspace{0.5em}

\begin{abstract}
Cardiovascular disease risk prediction models often rely on data from a single institution or centrally pooled datasets. Extending these models across institutions could be limited by privacy regulations and constraints on sharing patient-level data. Federated learning enables collaborative model development without transferring sensitive patient data, but its application in healthcare remains challenging because datasets often differ in size, population characteristics, and outcome definitions. In this study, we present a federated deep learning approach for privacy-preserving cardiovascular disease risk prediction that integrates two population-based cohorts with different characteristics: Lifelines, including 148,230 participants meeting the study inclusion criteria with self-reported outcomes, and the Rotterdam Study, including a smaller cohort of 10,155 participants with digitally linked clinical outcomes. Model performance was primarily evaluated on the Rotterdam Study because of its complete follow-up. Deep survival models trained using federated learning achieved higher predictive performance than models trained locally without federation. For the Rotterdam Study, the C-statistic increased from 0.728 (95$\%$ CI: 0.717–0.739) to 0.739 (95$\%$ CI: 0.728–0.749). For Lifelines, the C-statistic increased from 0.783 (95$\%$ CI: 0.775–0.791) to 0.787 (95$\%$ CI: 0.780–0.792). These findings suggest that federated deep learning across heterogeneous cohorts can improve cardiovascular disease risk prediction while preserving the privacy of individual-level patient data

\vspace{1em}

\end{abstract}

\section{Introduction}\label{sec:intro}
In healthcare, patient records contain sensitive health information that is classified as special-category personal data under regulatory frameworks such as the General Data Protection Regulation (GDPR), which restricts the sharing of personal data across organizational boundaries \citep{kaissis2021federated}. Building predictive models that draw on data from multiple institutions is therefore challenging, as pooling individual-level records across sites is often not permitted. Federated learning addresses this challenge by distributing the training process across several nodes, each holding a separate part of the data \citep{mcmahan2017communication,kaissis2021federated}. Rather than sharing raw records, each node trains a model on its local data and sends only the resulting model parameter updates, such as learned network weights, to a central server, which aggregates them into a global model 
\citep{mcmahan2017communication}. Since raw data never leave their source, federated learning has gained attention as a practical means to build predictive models across multiple institutions while protecting data privacy \citep{kaissis2021federated,li2025challenges}. \par
One domain where this capability is particularly valuable is cardiovascular disease (CVD) risk prediction. CVD remains a leading cause of morbidity and mortality worldwide \citep{eckel20142013,bakhit_cardiovascular_2024}, and accurate risk prediction plays a central role in primary prevention by identifying high-risk individuals before clinical events occur \citep{eckel20142013,bakhit_cardiovascular_2024}. Widely used risk prediction tools, such as the Pooled Cohort Equations \citep{goff20142013}, QRISK3 \citep{hippisley2017development}, and SCORE2 \citep{esc2021score2}, have been developed using data from a single institution or centrally pooled datasets \citep{alaa2019cardiovascular}. These tools are used in clinical practice to estimate an individual's 10-year risk of CVD events and to guide decisions on initiating preventive therapies such as statins or blood pressure-lowering medication \cite{arnett20192019,visseren20222021,mach20252025}. This limits the diversity of data available for model training and raises challenges when extending such models to new populations or settings, particularly in contexts where data sharing across institutions is restricted by privacy regulations. Federated learning offers a way to train CVD risk prediction models across multiple cohorts without pooling individual-level data, which may improve model generalizability while preserving privacy. \par
The established risk prediction tools are based on Cox proportional hazards regression. Deep survival models, such as DeepSurv \citep{faraggi1995neural,katzman2018deepsurv}, extend conventional Cox regression by capturing non-linear associations between risk factors and time-to-event outcomes, and have shown improved performance over Cox models for CVD risk prediction in independent cohorts \citep{hathaway2021deep,sasagawa2024application}. To investigate the feasibility of training such deep learning-based CVD risk prediction models in a federated setting, we previously introduced MyDigiTwin \citep{cadavid2025mydigitwin}, a privacy-preserving framework for personalized CVD risk prediction built on the Vantage6 federated learning platform \citep{moncada2020vantage6}. The framework includes a data harmonization pipeline that converts cohort data into an HL7 FHIR-compliant format, enabling consistent predictor extraction across participating sites. Technical feasibility of the infrastructure was previously demonstrated using the Lifelines cohort \citep{scholtens2015cohort}, where the harmonized dataset was partitioned across multiple simulated Vantage6 nodes to evaluate data distribution, local model updates, and aggregation. However, the study primarily focused on validating the infrastructure and workflow rather than evaluating the predictive benefit of true multi-site federated learning. Extending the framework to multiple sites introduces additional challenges because participating datasets may differ in size and population characteristics. Furthermore, differences in outcome definitions and follow-up information may affect model performance. \par
The present study extends the MyDigiTwin proof of concept to a real-world federated setting involving two population-based cohorts: Lifelines \citep{scholtens2015cohort} and the Rotterdam Study \citep{hofman2007rotterdam}. In contrast to the previous infrastructure-focused evaluation using a pseudo-split from a single cohort, this work investigates whether federated deep learning can improve cardiovascular risk prediction across different cohorts without exchanging individual-level participant data. By combining a large cohort with self-reported cardiovascular outcomes (Lifelines) and a smaller cohort with digitally linked clinical outcomes (Rotterdam Study), this study provides a practical evaluation of federated learning for CVD risk prediction in a heterogeneous multi-cohort setting. \par

\section{Materials and methods}\label{sec:methods}

\subsection{Cohort study}\label{ssec:populations}

\subsubsection{Lifelines}\label{sssec:lifelines}
The Lifelines Cohort Study, established in 2006, is a large population-based prospective study and biobank designed to enhance our understanding of healthy aging among residents of the northern part of the Netherlands \citep{scholtens2015cohort,van2017lifelines}. Recruitment took place between 2006 and 2013, primarily through general practitioners, with additional participants enrolled through family members of existing participants or self-registration, resulting in a three-generation cohort of 167,729 participants. Participants undergo repeated assessments, including a baseline questionnaire and physical examination, with follow-up questionnaires and visits. Self-reported questionnaires cover demographics, lifestyle, and health status, while physical examinations and biomaterial collection provide additional clinical measurements. For the present study, participants were included if they had no prior CVD event recorded before the start of follow-up, resulting in 148,230 participants eligible for analysis. \par

\subsubsection{Rotterdam Study}\label{sssec:rs}
The Rotterdam Study is an ongoing population-based prospective cohort study initiated in 1990 in the Ommoord district of Rotterdam, the Netherlands \citep{hofman2007rotterdam,ikram2024rotterdam}. The study was established to investigate the causes, development, and prognosis of chronic diseases in older adults. For the present study, participants were selected from the third examination round of the first cohort (RS-I-3) and the first examination rounds of the second and third cohorts (RS-II-1 and RS-III-1). Individuals who provided informed consent for follow-up were eligible for inclusion. Participants undergo assessments, including physical examinations, laboratory measurements, imaging examinations, and continuous follow-up through linkage with healthcare records. Individuals who provided informed consent for follow-up and had no prior CVD event recorded before the index date were eligible for inclusion. After applying these criteria and retaining participants with available predictor data after harmonization, 10,155 participants were included in the analysis. \par

\subsection{Definitions and Data harmonization}\label{ssec:data}

\subsubsection{Predictor selection}\label{sssec:predictors}
Predictors were selected based on their reported associations with CVD risk and their use in widely adopted CVD risk prediction tools, including the Pooled Cohort Equations, QRISK3, and SCORE2. The selection covers key risk domains: demographic characteristics, blood pressure, lipid biomarkers, kidney function markers, smoking status, and prevalent cardiometabolic conditions. The final predictor set was additionally constrained by the variables consistently available in both cohorts. As shown in Table \ref{tab:characteristics}, the final set of 12 predictors comprised sex, age, systolic blood pressure (mmHg), diastolic blood pressure (mmHg), HDL cholesterol (mmol/L), LDL cholesterol (mmol/L), total cholesterol (mmol/L), estimated glomerular filtration rate (eGFR), creatinine ($\mu$mol/L), smoking status (never smoked, ex-smoker, current smoker), prevalent type 2 diabetes, and prevalent hypertension. \par

\subsubsection{Outcome definition}\label{sssec:outcome}
The outcome was a composite CVD endpoint defined as the first occurrence of stroke, myocardial infarction (MI), or heart failure (HF) during a 10-year follow-up period, following the same definitions used in the MyDigiTwin proof of concept \citep{cadavid2025mydigitwin}. In the Rotterdam Study, events were identified through continuous digital linkage to hospital discharge records and general practitioner files, providing exact event dates and complete follow-up. In Lifelines, events were derived from self-reported questionnaire data collected at repeated assessment waves. Participants who reported a CVD event at baseline were excluded as prevalent cases. For each event type (stroke, MI, and HF), incident events occurring during follow-up were identified from repeated assessment waves. For events first reported at a follow-up assessment wave, the event date was approximated as the midpoint between the date of the last event-free assessment and the date of the assessment at which the event was first reported. This interval-based approximation, specific to the Lifelines cohort, introduces imprecision in event timing that is inherent to self-reported longitudinal data. \par

\begin{table}[t!]
\caption{Participant characteristics in the Lifelines and Rotterdam
Study cohorts included in this study.}
\label{tab:characteristics}
\centering

\begin{tabular}{llcc}
  \toprule
  \multirow{2}{*}{Clinical characteristic} &
  \multirow{2}{*}{Description} &
  \multicolumn{2}{c}{Cohort} \\
  \cmidrule(lr){3-4}
  & & Lifelines & Rotterdam Study \\
  \midrule
  Total number of individuals    & $N$              & $148{,}230$           & $10{,}155$            \\
  Women                          & $N$ (\%)         & $87{,}381$ $(58.9)$   & $6{,}114$ $(60.2)$    \\
  CVD event during follow-up     & $N$ (\%)         & $3{,}907$ $(2.6)$     & $1{,}338$ $(13.2)$    \\
  Age, years                     & Median (Q1; Q3)  & $44.2$ $(35.4; 51.3)$ & $62.7$ $(57.6; 71.5)$ \\
  Systolic blood pressure, mmHg  & Mean (SD)        & $127.1$ $(16.1)$      & $139.3$ $(21.1)$      \\
  Diastolic blood pressure, mmHg & Mean (SD)        & $73.7$ $(9.5)$        & $79.2$ $(11.4)$       \\
  HDL-cholesterol, mmol/L        & Mean (SD)        & $1.5$ $(0.4)$         & $1.4$ $(0.4)$         \\
  LDL-cholesterol, mmol/L        & Mean (SD)        & $3.3$ $(0.9)$         & $3.7$ $(0.9)$         \\
  Total cholesterol, mmol/L      & Mean (SD)        & $5.1$ $(1.0)$         & $5.8$ $(1.0)$         \\
  eGFR, ml/min/1.73\,m$^{2}$    & Mean (SD)        & $91.2$ $(16.0)$       & $85.0$ $(14.2)$       \\
  Creatinine, $\mu$mol/L         & Mean (SD)        & $76.2$ $(14.7)$       & $76.9$ $(17.2)$       \\
  Never smoked             & $N$ (\%)         & $65{,}335$ $(45.8)$   & $3{,}207$ $(32.0)$    \\
  Ex-smoker                & $N$ (\%)         & $51{,}986$ $(36.5)$   & $4{,}481$ $(44.7)$    \\
  Current smoker           & $N$ (\%)         & $25{,}273$ $(17.7)$   & $2{,}336$ $(23.3)$    \\
  Type 2 diabetes                & $N$ (\%)         & $3{,}998$ $(2.7)$     & $146$ $(1.4)$         \\
  Hypertension                   & $N$ (\%)         & $36{,}557$ $(24.7)$   & $3{,}720$ $(36.6)$    \\
  \bottomrule
\end{tabular}
\end{table}

\subsubsection{Data harmonization pipeline}\label{sssec:harmonization}
To enable federated learning across the two cohorts, both datasets were transformed into a standardized, HL7 FHIR-compliant format using the MyDigiTwin data harmonization pipeline \citep{cadavid2025mydigitwin}. The pipeline was developed to address the structural and semantic differences between cohort datasets that use different variable names and coding systems, which would otherwise prevent a federated algorithm from operating on consistent inputs across sites. \par
The harmonization process follows a three-stage workflow. First, raw tabular cohort data in CSV format were converted into patient-centered Common Data Format (CDF) files, which organize
each participant's data as a longitudinal record structured around assessment timepoints. Second, dataset-specific mapping rules, referred to as pairing rules, were applied to transform each CDF file into FHIR resources compliant with the ZIB profile used in Dutch personal health environments \citep{cadavid2024leveraging}. These pairing rules encode the logic required to map each source
variable to the correct FHIR element. Third, the resulting FHIR resources were flattened into a tabular structure that could be used by the federated learning algorithm. Both the Lifelines and Rotterdam Study datasets were harmonized independently at their respective sites using this pipeline; no raw data were transferred between sites at any stage of this process.

\subsection{Federated learning setup}\label{ssec:fl}

\subsubsection{Risk prediction model}\label{sssec:deepsurv}
DeepSurv \citep{katzman2018deepsurv} was used as the CVD risk prediction model.
DeepSurv extends the Cox proportional hazards (PH) model \citep{cox1972regression} by replacing its linear predictor with a feed-forward neural network, allowing non-linear associations
between predictors and time-to-event outcomes to be learned from data. In the standard Cox PH model, the hazard function is expressed as  
\begin{equation}\label{eq:cox}
h(t, \bm{X}) = h_{0}(t) \exp(\eta = \bm{X}\bm{\beta}),
\end{equation}  
\noindent where $h_{0}(t)$ is the unspecified baseline hazard function, $\bm{X} \in \mathbb{R}^{1 \times P}$ is the predictor vector with $P$ predictors, $\bm{\beta}$ are the regression coefficients, and $\eta$ is the linear log-risk score. In DeepSurv, the linear term $\bm{X}\bm{\beta}$ is replaced by $h_{\theta}(\bm{X})$, the output of a feed-forward neural network
with parameters $\theta$:  
\begin{equation}\label{eq:cox_deepsurv}
h(t, \bm{X}) = h_{0}(t) \exp(\eta = h_{\theta}(\bm{X})).
\end{equation} 
\noindent The network is trained by minimizing the negative partial log-likelihood \citep{faraggi1995neural}: 
\begin{equation}\label{eq:log_likelihood}
LL(\theta) = -\frac{1}{\sum_i \delta_i}
\sum_{i=1}^{N} \delta_i
\left\{
  h_{\theta}(\bm{X}_i)
  - \log\!\left[\sum_{j \in R_i}
      \exp\!\left(h_{\theta}(\bm{X}_j)\right)\right]
\right\},
\end{equation} 
\noindent where $\delta_i \in \{0,1\}$ is the event indicator, $N$ is the batch size, $R_i$ is the risk set of individuals who have not yet experienced the event at time $t_i$, and $h_{\theta}(\bm{X}_i)$ is the network output for individual $i$. Survival time was defined as the interval from cohort entry to the composite CVD endpoint described in \Cref{sssec:outcome}, with right-censored observations included through the risk set $R_i$. \par
The network architecture comprised an input layer of $P = 12$ neurons corresponding to the 12 predictors described in \Cref{sssec:predictors}, two hidden layers of 16 neurons each, and
a single linear output neuron, denoted $[12, 16, 16, 1]$. ReLU activation and batch normalization were applied after each hidden layer. A dropout rate of 0.20 was applied before each hidden layer. 
Model parameters were optimized using the Adam optimizer \citep{kingma2014adam} with a mini-batch size of 1024.
The learning rate was set to $1 \times 10^{-4}$, and each local model was trained for 100 epochs per federated round. A fixed random seed was used across all experiments to ensure
reproducibility. \par

\subsubsection{Federated learning algorithm}\label{sssec:fl_algm}
\begin{figure}[th]
	\centering
	\includegraphics[width=0.6\linewidth]{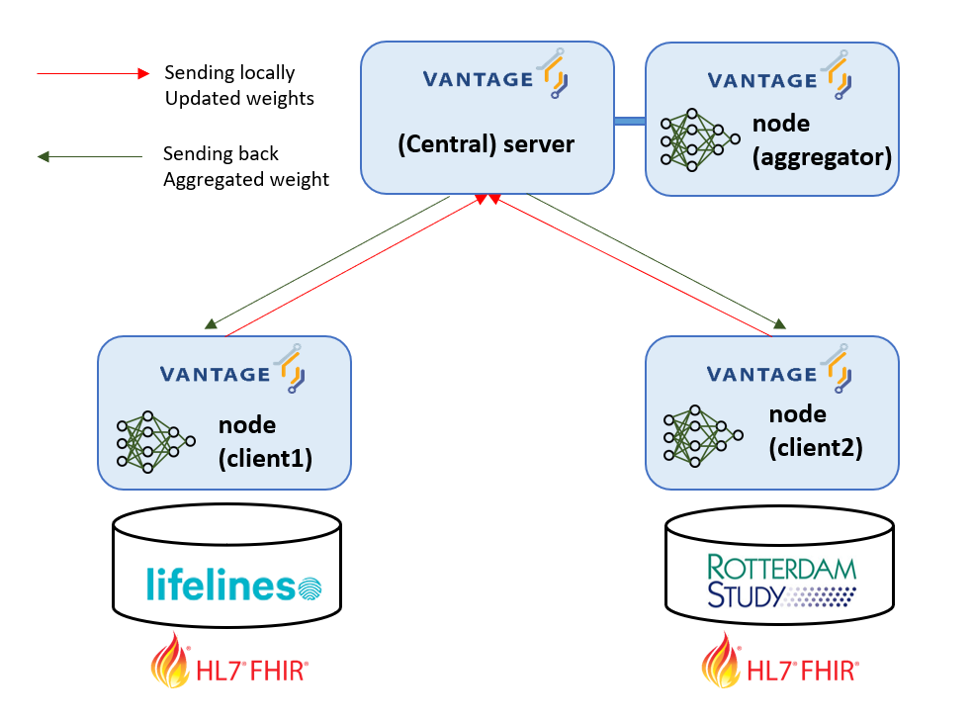}
	\caption{Overview of the federated learning setup using the Vantage6
infrastructure. A dedicated aggregator node receives locally updated
model weights from two data nodes (Lifelines and Rotterdam Study) and
broadcasts the aggregated weights back, without exchanging
individual-level data.}
	\label{fig:fl_arch_overview}
\end{figure}

Because each site holds a distinct set of participants with the same feature space, the data across sites are horizontally partitioned \citep{kairouz2021advances}, and horizontal federated
learning was applied. The federated setup follows a client-server architecture implemented
on the Vantage6 platform \citep{moncada2020vantage6}, as illustrated in \Cref{fig:fl_arch_overview}. Two data nodes (client 1: Lifelines; client 2: Rotterdam Study) each
hold a locally stored, FHIR-harmonized dataset and perform local model training independently.
A separate aggregator node, connected to the central Vantage6 server, receives the locally updated model weights from all client nodes and computes the aggregated weight, which is then broadcast back to the clients for the next round. No individual-level data leave the local environment at any point. \par
Model aggregation followed Federated Averaging (FedAvg) \citep{mcmahan2017communication}, in which the aggregated weight $\overline{w}_{t+1}$ is computed as the weighted average of the locally updated weights $w^{k}_{t+1}$ across all $K$ client nodes:
 
\begin{equation}\label{eq:fedavg}
\overline{w}_{t+1} = \sum_{k=1}^{K} \frac{n_k}{n} w^{k}_{t+1},
\end{equation}
 
\noindent where $n_k$ is the number of training samples at node $k$ and $n = \sum_{k=1}^{K} n_k$ is the total number of training samples across all nodes. At the start of training, the aggregator initializes the model weights and broadcasts them to all client nodes. In each of the $I = 20$ aggregation iterations, every client node updates the received weights by local training on its own data and returns the updated weights to the aggregator. The aggregator then computes $\overline{w}_{t+1}$ according to \Cref{eq:fedavg} and broadcasts it to all clients for the next
round. Although more advanced aggregation strategies exist for non-IID settings, empirical evaluations of federated learning challenges have shown that FedAvg can perform competitively with more complex algorithms \citep{schmidt2024fair}, supporting its use as the aggregation method in this study.

\subsubsection{Evaluation metrics and experiment setting}\label{sssec:eval}

The Lifelines data node was hosted on a virtual machine running on UMCG's Azure Cloud infrastructure. The Rotterdam Study data node was hosted within the SURF Research Cloud, where pseudonymized data are stored and processed in a secure workspace. The central Vantage6 server and the aggregator node were hosted on separate, independent workspaces within the SURF Research Cloud, distinct from the Rotterdam Study data workspace. Data processing and model training were conducted under the governance framework established within the MyDigiTwin project, including a Joint Controller Agreement between participating institutions and a Data Protection Impact Assessment to ensure compliance with GDPR requirements. \par
At each data node, missing predictor values were imputed independently using median imputation. The local dataset was then partitioned into 10 independent runs using stratified resampling, with event status used as the stratification variable to preserve the event rate across splits.
In each run, 80\% of participants were used for training, 10\% for validation during local training, and 10\% as the held-out test set. For Lifelines ($N = 148{,}230$), this corresponded to 118,584 training and 14,823 test participants per run. For the Rotterdam Study ($N = 10{,}155$), this corresponded to 8,124 training and 1,016 test participants per run. Data splitting was performed independently at each local node. \par
Model discrimination was evaluated using the concordance statistic (C-statistic), with performance reported as the mean and 95\% confidence interval (CI) across the 10 independent runs.
CIs were computed using the corrected resampled $t$-test \citep{nadeau1999inference}, which adjusts for the statistical dependence between independent runs.  \par

\section{Results}\label{sec:results}

\begin{table}[t!]
\begin{center}
\caption{C-statistic with 95\% confidence intervals for the local model (without aggregation) and the federated model (FedAvg after 20 update iterations), estimated using a corrected resampled $t$-test over 10 independent runs.}\label{tab:lifelines_results}
\begin{tabular}{l>{\centering}p{4.8cm}>{\centering\arraybackslash}p{4.8cm}}
\toprule[1pt] 
\multirow{2}*{Model evaluation} & \multicolumn{2}{c}{C statistic (95\% CI)} \\
& Without aggregation & \makecell{FedAvg\\(after 20 updates)} \\
\midrule[0.5pt]
Rotterdam Study    & 0.728 (0.717-0.739) & 0.739 (0.728-0.749) \\
Lifelines          & 0.783 (0.775-0.791) & 0.787 (0.780-0.792) \\
\arrayrulecolor{black}\bottomrule[1pt]
\noalign{\vskip 0.3mm} 
\end{tabular}
\end{center}
\end{table}

Performance was compared between the local model, trained on each cohort independently without federation, and the federated model after 20 aggregation iterations of FedAvg. The \textit{without aggregation} setting represents the local DeepSurv model trained using only the data available at each individual data node, before any model weight aggregation. The \textit{FedAvg after 20 updates} setting represents the final federated model after 20 iterations of local training followed by central aggregation of model weights across the Lifelines and Rotterdam Study nodes. \par
\Cref{tab:lifelines_results} summarizes the discrimination performance of the local and federated models. Federated training improved the C-statistic in both cohorts compared with local training. In the Rotterdam Study, the C-statistic increased from 0.728 (95\% CI: 0.717--0.739) without aggregation to 0.739 (95\% CI: 0.728--0.749) with federated training. In Lifelines, the C-statistic increased from 0.783 (95\% CI: 0.775--0.791) without aggregation to 0.787 (95\% CI: 0.780--0.792) with federated training. For the Rotterdam Study, the point estimate of the federated model (0.739) coincided with the upper bound of the local model CI, while the point estimate of the local model (0.728) coincided with the lower bound of the federated model CI. Although the confidence intervals overlapped, the point estimates consistently favored federated training in both cohorts. \par
\begin{figure}[htbp]
\centering

\begin{minipage}{0.45\textwidth}
    \centering
    \includegraphics[width=\linewidth]{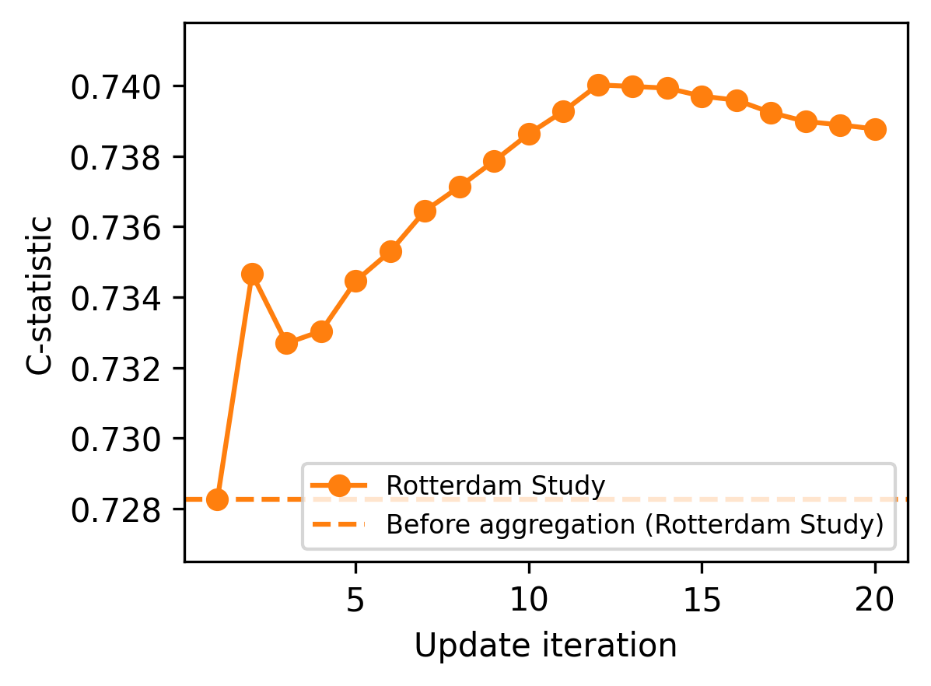}

    \vspace{2mm}
    \text{(A)}
\end{minipage}
\hfill
\begin{minipage}{0.45\textwidth}
    \centering
    \includegraphics[width=\linewidth]{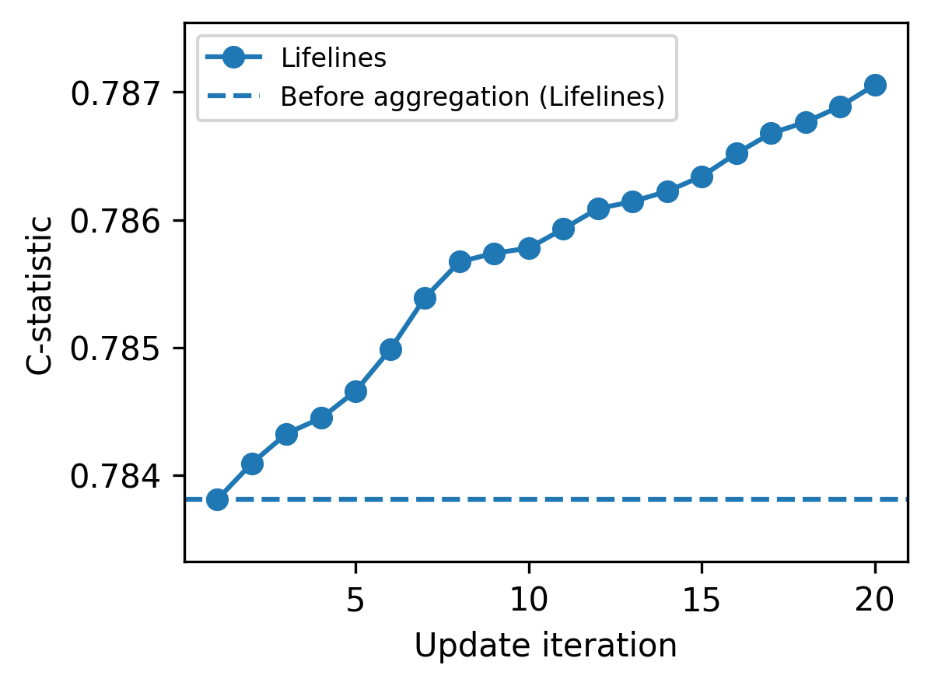}

    \vspace{2mm}
    \text{(B)}
\end{minipage}

\caption{Mean C-statistic across 10 independent runs at each FedAvg update iteration for the Rotterdam Study (A) and Lifelines (B). The dashed line indicates the C-statistic of the local model trained without federation. The solid line shows the C-statistic of the federated model at each update iteration.}
\label{fig:result_plot}

\end{figure}

Figure \ref{fig:result_plot}(A) shows the C-statistic across federated update iterations for the Rotterdam Study. Following the first update iteration, the C-statistic increased above the local baseline and, despite some fluctuations, reached a peak of approximately 0.740 around iteration 12 before showing a small decrease toward the end of training. Nevertheless, performance after 20 update iterations remained above the local model performance. \par
In Lifelines, Figure \ref{fig:result_plot} (B), the C-statistic increased more gradually across the 20 update iterations. The performance after the first update iteration was close to the local baseline of approximately 0.784, and subsequent update iterations showed
a steady, near-monotonic upward trend. Unlike the Rotterdam Study, no decrease was observed toward the later update iterations. No clear performance plateau was reached within the 20 update iterations evaluated, and the final model showed a higher C-statistic than the local model without aggregation. \par 
Overall, the results show that federated aggregation improved model discrimination in both cohorts at every update iteration after the first. The magnitude of improvement was larger in the Rotterdam Study than in Lifelines, while the trajectory of improvement differed between cohorts: the Rotterdam Study showed an early increase followed by a plateau and slight decline, whereas Lifelines showed a more gradual and sustained increase across all update iterations. \par

\section{Discussion}\label{sec:discussion}

In this study, we evaluated federated learning of deep survival neural networks for CVD risk prediction across two population-based cohorts with different characteristics. A DeepSurv model was trained using FedAvg between the Lifelines cohort and the Rotterdam Study without exchanging individual-level participant data. Compared with local training without federation, federated learning improved discrimination in both cohorts, with the C-statistic increasing from 0.728 to 0.739 in the Rotterdam Study and from 0.783 to 0.787 in Lifelines. The largest increase in mean C-statistic was observed in the Rotterdam Study, which served as the primary evaluation cohort because of its complete follow-up and digitally linked clinical outcomes. The larger improvement observed in the Rotterdam Study compared with Lifelines may be related to its smaller sample size and lower baseline discrimination performance prior to federation, allowing it to benefit more from information contributed by the substantially larger Lifelines cohort.\par
The present study extends earlier infrastructure-focused evaluations of federated learning for CVD risk prediction by assessing whether federated deep learning of CVD risk prediction can improve
performance across two independent population-based cohorts that differ in sample size, population characteristics, outcome collection procedures, and follow-up completeness. In real-world federated learning applications, such heterogeneity can make collaborative model development challenging because local models are optimized on data collected under different conditions \citep{mateus2025multi}. Despite these differences, federated training improved discrimination in both cohorts, demonstrating that collaborative model training remained beneficial in a heterogeneous multi-cohort setting. \par
A key question motivating this work was whether a large cohort with less precise outcome information could contribute to improving risk prediction when combined with a smaller cohort with precise, digitally linked outcomes. In Lifelines, cardiovascular events were derived from self-reported questionnaires and event dates were approximated from assessment intervals, meaning both the event indicator and event timing are subject to greater uncertainty than in the Rotterdam Study. Nevertheless, incorporating Lifelines through federated learning improved performance on the Rotterdam Study, suggesting that the information contained within the substantially larger Lifelines cohort may have compensated, at least partially, for the uncertainty associated with self-reported outcomes and interval-based event timing. The present findings suggest that such cohorts may still provide value within a federated learning framework when combined with smaller cohorts containing more accurate outcome data. \par
The convergence trajectories across update iterations further illustrate the differences between a real-world federated setting and a pseudo-split setting based on a single cohort. In the previous proof-of-concept study \citep{cadavid2025mydigitwin} using a pseudo-split Lifelines dataset, model performance increased monotonically across aggregation iterations. In contrast, the present study involved two independent cohorts with different population characteristics, event rates, and outcome definitions. The Rotterdam Study showed a rapid improvement during the early update iterations, with performance peaking around iteration 12 before a slight decline, whereas Lifelines showed a more gradual increase throughout training. One possible explanation for these differing trajectories is the substantial imbalance in cohort size between the participating sites. Because FedAvg weights local updates according to the number of training samples at each site, the substantially larger Lifelines cohort contributes more strongly to the aggregated model than the Rotterdam Study. Consequently, the influence of the Lifelines model may increase with successive aggregation rounds, potentially contributing to the slight decline in Rotterdam Study performance observed during later update iterations. Nevertheless, both cohorts achieved higher discrimination after federated training than before aggregation, suggesting that information learned from one cohort was transferable to the other despite their differences. These findings highlight the importance of monitoring convergence behavior in real-world federated settings and suggest that adaptive stopping criteria may be beneficial when training across heterogeneous cohorts. \par
From a privacy and governance perspective, the federated setup is consistent with the requirements of the GDPR and the data governance agreements in place for both cohorts. Only aggregated model weights were shared between nodes, and no individual-level data left the local environment at any point. Pseudonymized Rotterdam Study data were processed exclusively within the SURF Research Cloud, while Lifelines data remained within a dedicated virtual machine on UMCG's Azure Cloud infrastructure, with algorithm authorization controlled by the Lifelines data administrator. These findings demonstrate that meaningful multi-cohort federated deep learning for CVD risk prediction is achievable within existing institutional privacy frameworks without requiring any relaxation of data protection requirements. \par
Several limitations of this study should be noted. The federated setup involved only two data nodes, which limits the generalizability of the results to settings with more participating institutions. The Lifelines outcome labels carry inherent imprecision due to their retrospective self-reported nature and interval-based event timing, which may have limited the contribution of Lifelines to the federated model. In addition, the relatively high age of participants in the Rotterdam Study implies a substantial risk of non-CVD mortality during follow-up. In the present analysis, deaths from causes other than CVD were treated as censoring events rather than competing risks, which may result in some overestimation of absolute CVD risk in older individuals. Future work should evaluate competing-risk survival models that explicitly account for non-CVD mortality in older populations. Future work should extend this approach to additional cohorts and sites, and evaluate more robust aggregation strategies for non-IID data, including methods that address label distribution shift such as FedERFT \citep{zhu2026federft}, prototype-based clustering approaches such as FedACA \citep{dong2026fedaca}, and multi-objective aggregation frameworks that support different model architectures per client \citep{ma2026multi}, which may be beneficial given the large difference in sample size between the two cohorts. Alternative aggregation schemes that account for differences in cohort size or data quality, rather than relying solely on sample-size-based weighting as in FedAvg, may also be beneficial in settings where smaller cohorts contain more precise outcome information than larger cohorts. Incorporating differential privacy and secure aggregation mechanisms, and extending the framework to include time-variant predictors and longitudinal data, are additional important directions for advancing the clinical utility of federated deep learning for CVD risk prediction. Future work should also investigate the integration of imaging-derived biomarkers, such as coronary artery calcification measures extracted from cardiac CT, to enable multimodal federated risk prediction models that combine imaging data with traditional cardiovascular risk factors. Overall, the present findings support the feasibility of federated deep learning for collaborative CVD risk prediction across heterogeneous cohorts with differing population characteristics and outcome definitions while preserving data privacy, highlighting its potential for future large-scale multi-institutional studies.\par







\section*{Author Contributions}
H.M. was involved in the conception and design of the study, conducted the method development and experiments, contributed to the federated learning infrastructure setup and implementation, performed the data analysis, and wrote the manuscript. D.S. was involved in the conception and design of the study, contributed to the federated learning infrastructure setup and implementation, and reviewed the manuscript. M.A.B. contributed to the federated learning infrastructure setup for the Rotterdam Study data node and reviewed the manuscript. M.J.G.L. contributed to the analysis and interpretation of the Rotterdam Study data, and reviewed the manuscript. 
D.B. was involved in the method development, contributed to the analysis and interpretation of the Rotterdam Study data, and reviewed the manuscript. P.v.d.H. was involved in the conception and design of the study, contributed to the analysis and interpretation of the Lifelines data, and reviewed the manuscript. E.E.B. was involved in the conception and design of the study, conducted the method development, contributed to the analysis and interpretation of the federated learning results, and reviewed the manuscript.


\section*{Acknowledgments}
  This work is part of the project MyDigiTwin with project number 628.011.213 of the research programme "COMMIT2DATA - Big Data \& Health" which is partly financed by the Dutch Research Council (NWO). Furthermore, this work used the Dutch national e-infrastructure with the support of the SURF Cooperative using grant no. EINF-7675. The Rotterdam Study is funded through unrestricted research grants from Erasmus Medical Center and Erasmus University, Rotterdam, Netherlands Organization for the Health Research and Development (ZonMw).



\bibliographystyle{Frontiers-Harvard} 
\bibliography{ref}


\section*{Figure captions}

\end{document}